\documentclass[twoside,journal]{IEEEtran}
\usepackage{cite}
\usepackage{amsmath,amssymb,amsfonts}
\usepackage{algorithmicx}
\usepackage{algpseudocode}
\usepackage{multirow}
\usepackage{graphicx}
\usepackage[ruled,linesnumbered]{algorithm2e}
\usepackage{textcomp}
\usepackage{xcolor}
\usepackage{float}
\usepackage{soul}
\usepackage{bm}
\usepackage{subfigure}
\usepackage{multicol}
\usepackage{lipsum}
\usepackage{setspace}
\usepackage{url}
\usepackage[hidelinks]{hyperref}

\ifCLASSINFOpdf
\else
\fi
\makeatletter
\IEEEtriggercmd{\reset@font\normalfont\fontsize{7.0pt}{8.0pt}\selectfont}
\makeatother
\IEEEtriggeratref{1}

\begin{document}

%


\title{EV-Planner: Energy-Efficient Robot Navigation via Event-Based Physics-Guided Neuromorphic Planner}

\author{Sourav Sanyal (\IEEEmembership{Graduate Student Member, IEEE}),
            Rohan Kumar Manna,
        and Kaushik Roy (\IEEEmembership{Fellow, IEEE}) \\
\emph{Elmore Family School of Electrical and Computer Engineering, Purdue University}
\thanks{Manuscript received: July 20, 2023; Revised: November 9, 2023; Accepted: December 16, 2023.}
\thanks{This paper was recommended for publication by
Editor Tetsuya Ogata upon evaluation of the Associate Editor and Reviewers’ comments.}
\thanks{This work was supported in part by the Center for Brain-inspired Computing (C-BRIC), a DARPA sponsored JUMP center, the Semiconductor Research Corporation (SRC), the National Science Foundation, US Army Research Lab, and IARPA MicroE4AI. Sourav and Rohan contributed equally to implement this work. Code and video demo available at \href{https://github.com/souravsanyal06/EV-Planner }{{https://github.com/souravsanyal06/EV-Planner }}}
\thanks{Digital Object Identifier (DOI): see top of this page.}
}
\newcommand\blfootnote[1]{%
  \begingroup
  \renewcommand\thefootnote{}\footnote{#1}%
  \addtocounter{footnote}{-1}%
  \endgroup
}

\maketitle

\markboth{IEEE ROBOTICS AND AUTOMATION LETTERS. PREPRINT VERSION. ACCEPTED JANUARY, 2024}
{SANYAL \MakeLowercase{\textit{et al}.}: EV-PLANNER: Energy-Efficient Robot Navigation}

\begin{abstract}
Vision-based object tracking is an essential precursor to performing autonomous aerial navigation in order to avoid obstacles. Biologically inspired neuromorphic event cameras are emerging as a powerful alternative to frame-based cameras, due to their ability to asynchronously detect varying intensities (even in poor lighting conditions), high dynamic range, and robustness to motion blur. Spiking neural networks (SNNs) have gained traction for processing events asynchronously in an energy-efficient manner. On the other hand, physics-based artificial intelligence (AI) has gained prominence recently, as they enable embedding system knowledge via physical modeling inside traditional analog neural networks (ANNs). In this letter, we present an event-based physics-guided neuromorphic planner (EV-Planner) to perform obstacle avoidance using neuromorphic event cameras and physics-based AI. We consider the task of autonomous drone navigation where the mission is to detect moving gates and fly through them while avoiding a collision. We use event cameras to perform object detection using a shallow spiking neural network in an unsupervised fashion. Utilizing the physical equations of the brushless DC motors present in the drone rotors, we train a lightweight energy-aware {physics-guided neural network (PgNN)} with depth  inputs. This predicts the optimal flight time responsible for generating near-minimum energy paths. We spawn the drone in the Gazebo simulator and implement a sensor-fused vision-to-planning neuro-symbolic framework using Robot Operating System (ROS). Simulation results for safe collision-free flight trajectories are presented  with performance analysis, {ablation study} and potential future research directions. 
\end{abstract}

\begin{IEEEkeywords}
Event cameras, Neuromorphic vision, Physics-based AI, Spiking Neural Networks, Vision-based navigation
\end{IEEEkeywords}

%
\IEEEpeerreviewmaketitle

\section{Introduction}
\label{intro}
%
%
%
%

\IEEEPARstart{F}{or} performing vision-based robot navigation \cite{kak}, object tracking is a fundamental task. As navigation environments become increasingly challenging \cite{li2020autonomous} (due to increased demands on robot speed or to fly under reduced lighting conditions), the need for newer vision sensors arises. Biologically inspired \emph{event cameras} or Dynamic Vision Sensors (DVS) \cite{ev1, ev2, ev3, ev4}, which are capable of triggering \emph{events} in response to changes in the \emph{logarithm of pixel intensities past a certain threshold}, have emerged as a promising candidate. Event cameras are relatively immune to problems such as motion blur, can withstand higher temporal resolution ($10 \mu s$ vs $3 ms$), operating frequencies with wider dynamic illumination ranges ($140 dB$ vs $60 dB$), and consume lower power ($10 mW$ vs $3 W$), compared to traditional frame-based  cameras \cite{evsurvey}. Consequently, there has been considerable interest in the autonomous systems community, in using event cameras as vision sensors, for navigation purposes. Inspired by the neuronal dynamics observed in biological brains, spiking neurons, specifically \emph{Leaky Integrate and Fire} (LIF) neurons \cite{DELORME1999989} have been designed for leveraging temporal information processing -- quite similar to the signals produced by event cameras. Furthermore, the asynchronous event-driven nature of spiking neuron firing makes it a natural candidate for handling asynchronous events \cite{lee2020spike}.  A recent work \cite{nagaraj2023dotie} utilized a shallow spiking architecture for spatio-temporal event processing to perform low latency object detection for autonomous navigation systems.  

\begin{figure}[!t]
\begin{center}
   \includegraphics[width = 0.4\textwidth]
   {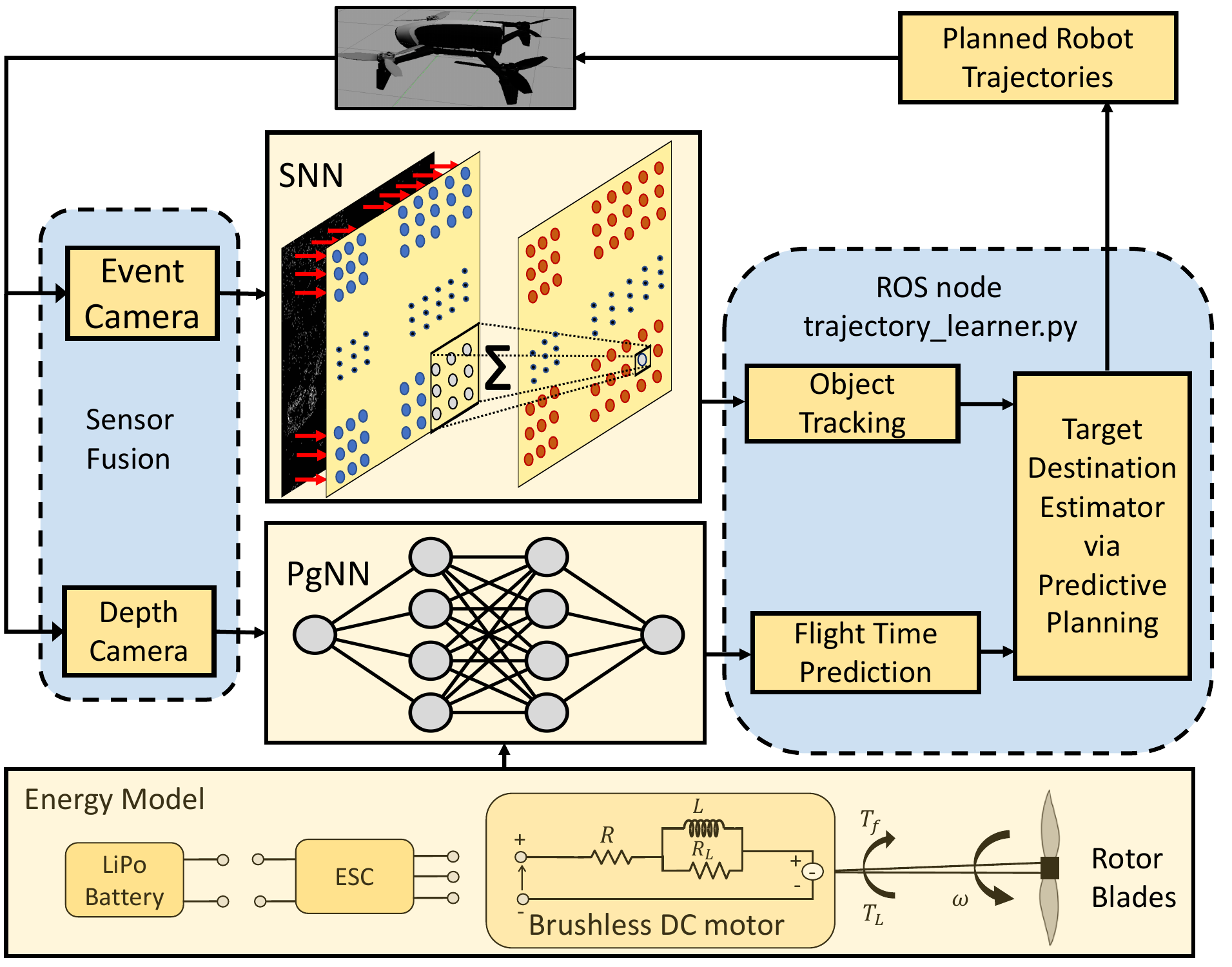} 
\end{center}
\vspace{-3mm}
    \caption{High-level overview of EV-Planner framework }
    \label{fig:method}
    \vspace{-5mm}
\end{figure}

 On the other hand, physics-based machine learning/artificial intelligence \cite{raissi2019physics, karniadakis2021physics, PINC, neuro-ising} is gaining momentum, as they enable encoding prior knowledge of physical systems while learning from data. In robotic systems, symbolic dynamics have been used traditionally to describe physical properties (of both systems and environments) \cite{4141034}. Works on predictive control \cite{NICODEMUS2022331, rampnet, salehi2023data, chee2022knode} have incorporated such physics-encoded prior inside neural network pipelines, and have shown to present advantages in terms of training efficiency (data and hence compute) as well as inference (latency and therefore energy). {Moreover, adding physical knowledge as priors reduces laborious dataset preparation efforts \mbox{\cite{wadekar2021towards}} and makes neural networks more robust and interpretable -- a quality immensely desired by the AI community in mission-critical scenarios.}

Considering today's actuators with a $15V$ supply, the average instantaneous  power for a simulated {Parrot Bebop2} is $\sim 124W$  (see Section \ref{phyro}), while the maximum hovering time without any maneuver is $\sim 25$ minutes for a $2700mAh$ battery \cite{parrot_link}.
For agile navigation tasks, the power increases, further reducing the flight time. To that effect,
we present \emph{EV-Planner -- an event-based physics-guided neuromorphic planner} to perform autonomous navigation energy-efficiently using neuromorphic event cameras and physics-based AI. Exploiting the energy-efficient nature of spike-based computation while efficiently capturing the temporal event information, and harnessing the benefits of physics-based simulations in training neural networks, the main goal of this work is to perform energy-efficient generalizable planning using a sensor fusion of event and depth cameras for {small drones.}  {We consider the task of autonomous quadrotor navigation where the mission is to detect moving gates and fly through them  without collisions.}

Figure \ref{fig:method} illustrates the logical overview of the EV-Planner framework. We consider two sensors -- an event camera and a depth camera. The Spiking Neural Network (SNN) block detects objects using event camera inputs. The Physics-guided Neural Network (PgNN) block learns to predict near-optimal trajectory times to the target destination from the depth input. The main contributions of this work are:
\begin{itemize}
    \item We use event cameras to perform object detection in a 3D environment using a shallow low-latency spiking neural architecture in an unsupervised fashion (Section \ref{perception}).
    \item Utilizing the physical equations of the brushless DC motors present in the drone rotors, we train a PgNN with depth camera inputs. The pre-trained network predicts near-optimal flight times responsible for generating near-minimum energy paths during inference (Section \ref{phyro}).
    \item We spawn a {Parrot Bebop2} drone in the Gazebo simulator and implement a sensor-fused vision-to-planning neuro-symbolic framework (Section \ref{symbol}) using the Robot Operating System (ROS) \cite{rotorS}.
    \item Using our simulation setup (Section \ref{method}), we perform autonomous collision-free trajectory planning. We present performance statistics {along with an ablation study to quantify the role of each design component} (Section \ref{res}).
\end{itemize}
\emph{To the best of our knowledge, this is the first work that uses event-based spike-driven neuromorphic vision coupled with physics-based AI for vision-based autonomous navigation.
This results in a hybrid neuro-symbolic network architecture for energy-efficient generalizable robot planning.}

\section{Background and Related Work}
We consider vision-based autonomous navigation systems which use event-based perception for obstacle avoidance and physics-guided robot learning for planning and control. Relevant research pursuits to achieve this are briefly discussed.

\subsection{\textbf{Event-based Robot Perception}}

Event cameras such as Dynamic Vision Sensors (DVS) capture asynchronous changes in the \emph{logarithm of light intensities}. Events are encoded as arrayed tuples of event instances, given as $E = \{ e_1, e_2, e_3, ... e_n \}$ across $n$ time-steps. Each discrete event $e_i$ $\forall$ $i=\{1,2,3, ..., n\}$ is represented as $e_i = \{x_i, y_i, p_i, t_i \}$, where $(x_i, y_i)$ is the location of the camera pixel, $p_i$ represents the \emph{polarity} (or sign-change of logarithm of intensity), and $t_i$ corresponds to $i$-th time-step.

Early efforts on event-based object detection applied simple kernels on event outputs to capture salient patterns \cite{ev-r1, ev-r3, ev-r4}. 
Although these works paved the way for future endeavors, they failed to capture events generated by the background. To overcome this, \cite{ev-r6} proposed a motion compensation scheme using ego-motion filtering to eliminate background events.
For object-tracking, \cite{ev-tracking-1} utilized both frames as well as event cameras. These hybrid techniques were computationally expensive as they relied on frame inputs to detect objects. Works in \cite{evreflex, evcatcher} estimate time-to-collision using events.
 Related to our work, event cameras have been mounted on drones for dynamic obstacle dodging \cite{scirobotics.aaz9712}. However, optical flow and segmentation using learning-based techniques require bulky neural networks with lots of parameters. In this work, we  focus on object detection for tracking using spiking neurons \cite{nagaraj2023dotie} -- achievable with low-latency neural networks. \\
 
\begin{table}[!t]
\label{tab:rw}
\centering
\tiny
\caption{Perception Works for 3D Object Tracking}
\vspace{-2mm}
\renewcommand{\arraystretch}{0.3}
\begin{tabular}{c|c}
\hline
\footnotesize{\textbf{Sensor Modality}} & \footnotesize{\textbf{Available Literature}} \\ \hline
\footnotesize{Events}                   & \footnotesize{\cite{nagaraj2023dotie, ev-r3, ev-r4,  ev-r6,  _EV3, _EV4}}                    \\ \hline
\footnotesize{Depth}                    & \footnotesize{\cite{D1, D2, D3, D4}}                 \\ \hline
\footnotesize{Events + Depth}          & \footnotesize{\emph{EV-Planner}}                 \\ \hline
\end{tabular}
\vspace{-6mm}
\end{table}
\raggedbottom

\noindent{\textbf{Why Sensor Fusion of Events and Depth?}}
  In order to track an object in a 3D environment, the notion of depth is of utmost importance. As shown in Table \ref{tab:rw}, there are works that either 1) estimate depth using non-depth sensors\cite{nagaraj2023dotie, ev-r3, ev-r4, ev-r6, _EV3, _EV4}  or 2) use depth sensors\cite{D1, D2, D3, D4}. The absence of an event sensor makes sensing and perception energy-expensive. On the other hand, the absence of a depth sensor will require estimating depth adding to the processing delay of the perception algorithm. Hence, in order to limit the overheads of perception delays (which is responsible for real-time object tracking), it is imperative that there are dedicated sensors for both events and depth. While adding sensors increases the  robot payload, future designs will consider embedding multiple sensors into a single device. The reduced  perception time will make the design more energy-efficient, rendering sensor fusion a promising design paradigm.

\subsection{\textbf{Physics-Based Robot Learning for Planning/Control}}
\label{pbl}
Physics-based AI shows promise as these approaches embed system knowledge via physical modeling inside neural networks, making them robust and interpretable.
{Traditionally, in robotic systems, symbolic methods have been used for describing physical properties as they are fast and accurate \mbox{\cite{4141034, rohan}}. 
For controlling dynamical systems, the work in \mbox{\cite{PINC}} added the provision of control variables to physics-informed neural networks. The work in \mbox{\cite{salehi2023data}} has showcased the advantages of using physical priors in control tasks. The research presented in \mbox{\cite{NICODEMUS2022331}} used \mbox{\cite{PINC}} for controlling multi-link manipulators, while the work in \mbox{\cite{ rampnet}} extended the previous framework for trajectory tracking of drones subjected to uncertain disturbances. The authors in \mbox{\cite{chee2022knode}} used neural ordinary differential equations (ODEs) for drones. While all these works use physics to solve differential equations, the same principle can be applied to add non-ODE (or non-PDEs) as soft constraints to neural network training, which is explored in this work. Incorporating physics into neural networks as a hard constraint still remains unexplored and is beyond the scope of this work.
}  

\section{EV-Planner}

  We present \emph{EV-Planner} -- an event-based physics-guided neuromorphic planner to perform obstacle avoidance using neuromorphic event cameras and physics-based AI. For object tracking, we use event-based low-latency SNNs and train a lightweight energy-aware neural network with depth inputs. 
 This makes our design loosely coupled with no interaction between the SNN and {PgNN block (see Figure \mbox{\ref{fig:method})}. Information about the current object position (via SNN block) and estimated trajectory time (via PgNN block) is consumed by a symbolic program (a ROS node$^2$)
 for predictive planning.}  

\def\thefootnote{2}\footnotetext{The basic unit of compute in ROS is called a node. A planner which performs decision-making via symbolic rules is implemented as a ROS node.}\def\thefootnote{\arabic{footnote}}
\subsection{\textbf{Event-based Object Tracking via Spiking Neural Network}}
\label{perception}
Due to a lack of photo-metric features such as light intensity and texture in event streams, traditional computer vision algorithms fail to work on events. Hence, we use a modified version of \cite{nagaraj2023dotie} which utilizes the inherent temporal information in the events with the help of a shallow architecture of spiking LIF neurons. The SNN can separate events based on the object's speed (present in the robot's field of view).

\begin{figure}[!t]
\begin{center}
   \includegraphics[width = 0.3\textwidth]
   {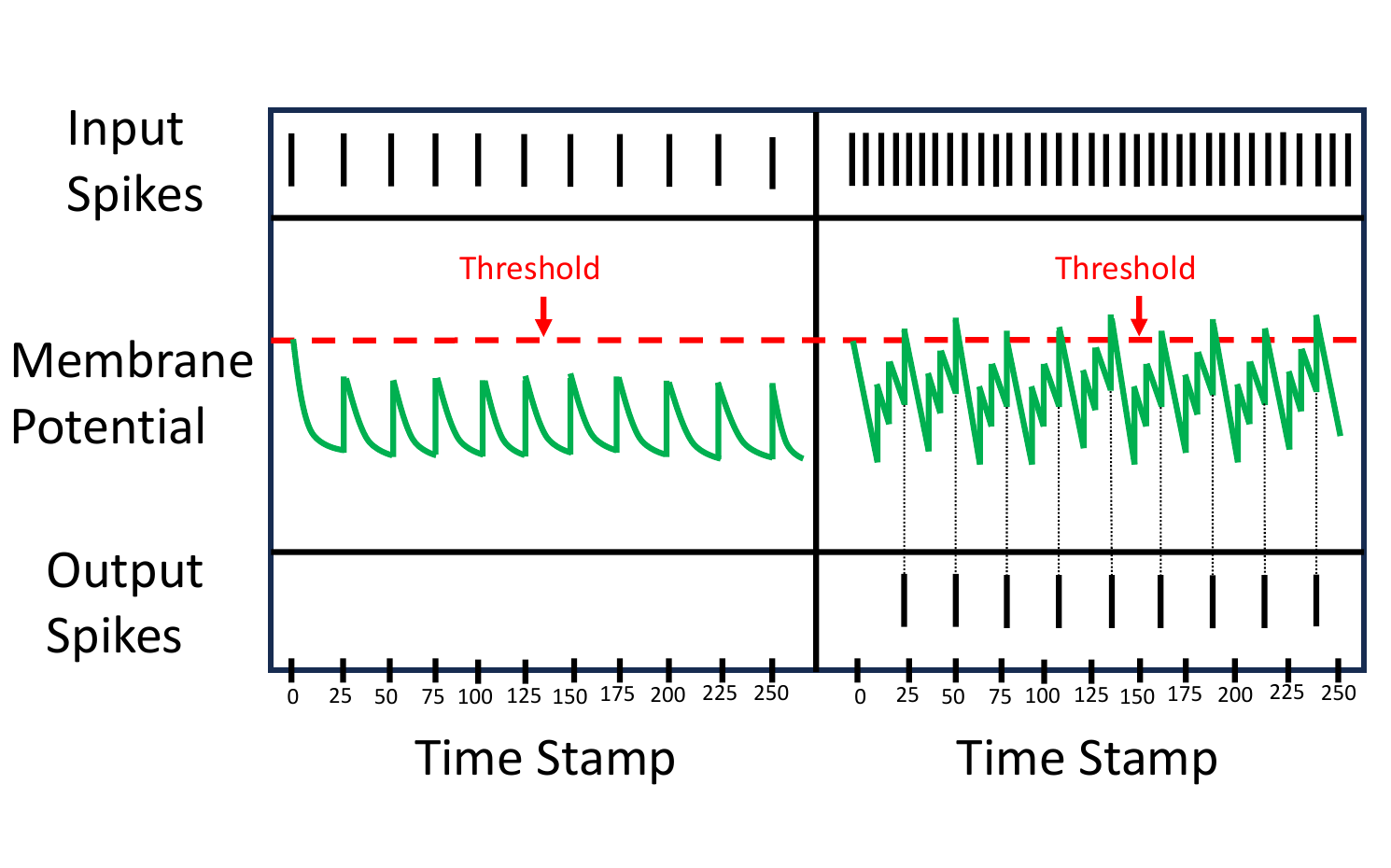} 
\end{center}
\vspace{-3mm}
    \caption{On the left, the spikes generated do not exceed the membrane potential threshold due to the absence of enough events. On the right, due to object motion, densely generated events produce more input spikes which exceed the membrane potential threshold to generate output spikes.}
    \label{fig:lif}
    \vspace{-3mm}
\end{figure}
\subsubsection{\textbf{Filtering out objects based on Speed}}
The network architecture used for filtering objects based on their motion consists of a single spike-based neural layer, whose parameters are fine-tuned to directly represent the speed of the moving object. 
The LIF neuron model \cite{DELORME1999989} has a membrane potential ($U[t]$) and a leak factor ($\beta$). At first, the initial membrane potential ($U[t_0]$) and the threshold value ($U_{th}$) get initialized. Next, according to equation [\ref{lif}] for each time step ($t$) the weighted sum of the inputs ($W X[t]$) gets stored in $U[t]$. Finally, if at any instance of time, the stored $U[t]$ exceeds $U_{th}$ then the neuron produces an output spike and resets $U[t]$.

\begin{equation} \label{lif}
U[t] = \beta U[t_{n-1}] + W X[t]
\end{equation}

In our approach hyper-parameters like $\beta$ and $U_{th}$ are fine-tuned to select nearby input spikes during any given instance of time. Figure \ref{fig:lif} demonstrates the difference in the output spikes generated from the neuron according to the frequency of the input events. 
As shown, the frequency of the generated events is directly proportional to the speed of the moving object. Exploiting this property of spiking neurons and event data, we consequently design an SNN capable of filtering out a particular object of interest, based on speed.

\begin{figure}[!t]
\begin{center}
   \includegraphics[width = 0.4\textwidth]
   {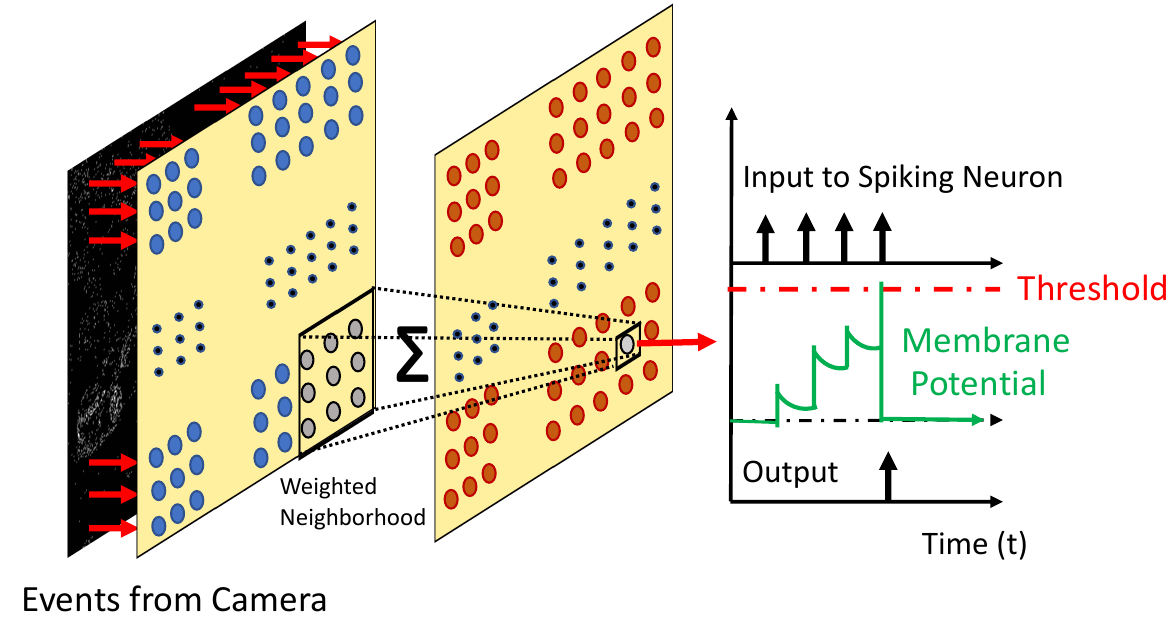} 
\end{center}
\vspace{-3mm}
    \caption{Spike-based neural architecture responsible for filtering events from objects based on speed. This SNN is used for subsequent object tracking.}
    \label{fig:nn_arch}
    \vspace{-3mm}
\end{figure}

\subsubsection{\textbf{Finding Center of the Bounding box}}
\label{find_centre}

To detect an object, the initial step involves connecting the output pixels from the event camera to the spiking neurons. Then, we apply a weighted sum ($W$) from a neighborhood of pixels  and pass that as input to the spiking neuron. Here, the neurons connected to the specific neighborhood  will register more than one input pixel every consecutive time step. This enables the SNN with the ability to identify pixels as fast-moving. In the next step, we locate the minimum ($X_{min}, Y_{min}$) and maximum ($X_{max}, Y_{max}$) pixel values in two dimensions and subsequently draw the bounding box. Finally, we obtain the center coordinates of the moving object using Eqn. (\ref{center}). 

\begin{subequations} \label{center}
    \begin{align}
    & center_x = X_{min} + \lfloor(X_{max}-X_{min})/2\rfloor \\
    & center_y = Y_{min} + \lfloor(Y_{max}-Y_{min})/2\rfloor
    \end{align}  
\end{subequations}

\begin{figure}[!t]
\begin{center}
   \includegraphics[width = 0.37\textwidth]
   {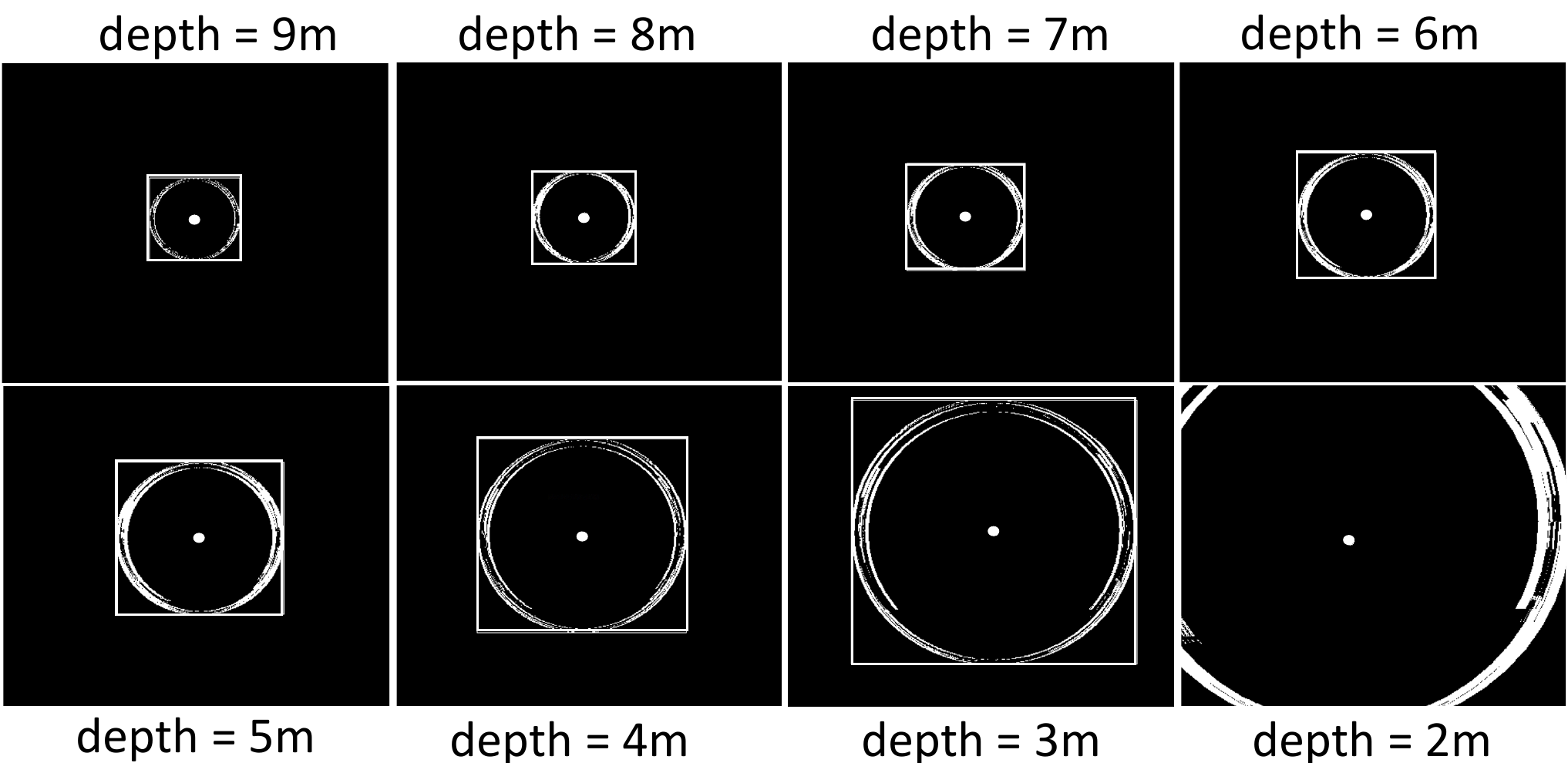} 
\end{center}
\vspace{-5mm}
    \caption{Isolating the moving object and finding the center of the bounding box in the pixel coordinate system at different depths.}
    \label{fig:center_box}
    \vspace{-1mm}
    \label{fig:evbox}
\end{figure}

Figure \ref{fig:nn_arch} depicts our network architecture. We selected a reasonable weighted neighborhood size ($W$) (see Section \ref{method}) as input to every spiking neuron. The weights for the neighborhood were normalized to obtain a summation value close to one. Note that the SNN mentioned above detects the moving object. But we still need to isolate the input events which represent the object of interest. Therefore, after detection, we recover the input events around the neighborhood of the spiking neuron output. Hence, the SNN works asynchronously to find the output spikes and recover the inputs from the previous time step (one-time step delay). Generally, after separating the objects based on the movement speed, their corresponding events are spatially grouped together using some clustering technique for predicting the object class. However, in this work, we only consider a single moving object. Hence, our current SNN-based design does not require any spatial clustering technique, which reduces the latency and energy overheads of our object detection and tracking. {Please note that the SNN is scene-independent and can be deployed in various scenarios with minimal fine-tuning.} Figure \ref{fig:evbox} illustrates the output of the event-based object (gate) tracking using SNN, situated at different depths (see Section \ref{track} for more details).

\vspace{-2mm}
\subsection{\textbf{Physics-Based Robot Modelling}}
\label{phyro}
\subsubsection{\textbf{Energy model}}
\begin{figure}[!t]
\vspace{2mm}
\begin{center}
   \includegraphics[width = 0.4\textwidth]
   {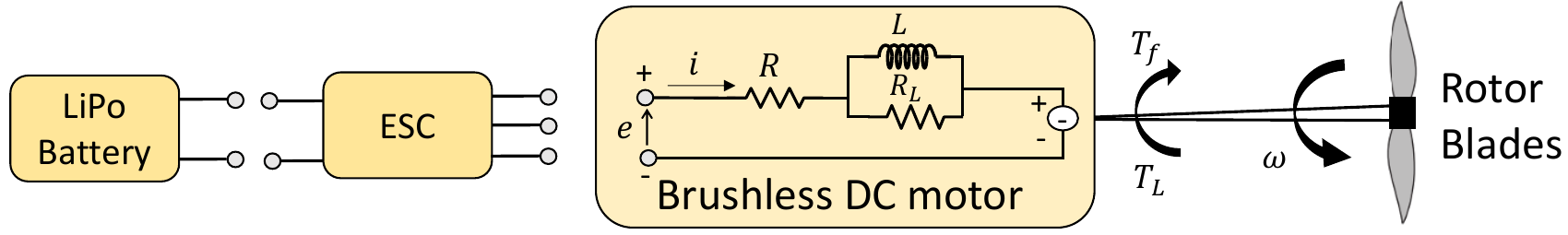} 
\end{center}
\vspace{-3mm}
    \caption{Energy model of brushless DC motor of quadrotor propellers}
    \label{fig:bldc}
    \vspace{-1mm}
\end{figure}

We model the actuation energy of the drone in flight by considering an electrical model of the LiPo battery-powered brushless DC (BLDC) motors of the rotor propellers \cite{energy}.
Figure \ref{fig:bldc} represents the equivalent circuit of each BLDC motor by considering the resistive and inductive windings. The energy for a particular flight path requires overcoming the motor and load frictions (given by the coefficients $J_m$ and $J_L$). The instantaneous motor current is:
\begin{equation} \label{current}
    i(t) = \frac{1}{K_T}[T_f + T_L \; \omega (t) + D_f \; \omega(t) + (J_m +J_L)\frac{d\omega (t)}{dt}]
\end{equation}

$K_T$ is the torque constant. $T_f$ and $T_L \; \omega (t)$ are the motor and load friction torques respectively. $D_f$ represents the viscous damping coefficient. For the BLDC motors of drone propellers, we can safely neglect $T_f$ and $D_f$. 
\begin{equation}
     J_L = \frac{1}{4}\; n_B \; m_B \; (r - \epsilon)^2 
\end{equation}
$n_B$ is the number of propeller blades, $m_B$ is the mass of the blade and $r$  is the propeller radius. $\epsilon$ is the clearance between the blade and motor.
\begin{table}[!t]
\tiny
\centering
\caption{Energy Model Specification}
\label{tab:en}
\renewcommand{\arraystretch}{0.3}
\begin{tabular}{c|c}
\hline
\footnotesize{\textbf{Parameters}}                       & \footnotesize{\textbf{Values}  }                \\ \hline
\footnotesize{Motor Resistance ($R$)}                    & \footnotesize{$0.3$ ohm  }                      \\
\footnotesize{Supply Voltage ($V$)    }                  & \footnotesize{$15$ V     }                      \\
\footnotesize{Maximum Motor Speed      }                 & \footnotesize{$7994$ rpm }                      \\
\footnotesize{Motor Friction Torque ($T_f$) }            & \footnotesize{$0.0187$ Nm }                     \\
\footnotesize{Aerodynamic Drag Coefficient ($k_{\tau}$)} & \footnotesize{$9.04969e-09$ }                   \\
\footnotesize{Viscous Damping Coefficient ($D_f$)  }     & \footnotesize{$2e-04$ Nm   }                    \\
\footnotesize{Motor Voltage Constant ($K_E = K_T$)   }         & \footnotesize{$0.532$ V/rpm  }                 \\
\footnotesize{Motor Moment of Inertia ($J_m$)  }         & \footnotesize{$4.9e-06$ kgm$^2$} \\
\footnotesize{Number of blades ($n_b$)}                  & \footnotesize{$3$}                                \\
\footnotesize{Mass of blade ($m_b$)  }                  & \footnotesize{$0.001$ kg}                         \\
\footnotesize{Radius of blade ($r$)   }                  & \footnotesize{$0.1$ m}                            \\
\footnotesize{Blade clearance ($\epsilon$) }             & \footnotesize{$0.023$ m}                          \\ \hline
\end{tabular}%
\end{table}
The motor voltage is as follows:
\begin{equation}
    e(t) = R\;i(t) + K_E\;\omega(t) + L\;\frac{di(t)}{dt}
\end{equation}
$R$ and $L$ denote the resistive and inductive impedances of the phase windings respectively. $K_E$ represents the motor voltage constant. We also neglect the electronic speed controller (ESC) voltage drops and any non-idealities present in the LiPo battery. For small drones (which we assume in this work), the propellers have direct connections with the shafts.
At steady-state operation, the motor voltage becomes:
\begin{equation} \label{voltage}
    e(t) = R\;i(t) + K_E \; \omega(t)
\end{equation}
For flight from time $0$ to $t$, then the total actuation energy of the flight path for all practical purposes can be written as:
\begin{equation} \label{energy}
\small
    E(t) = \int_0^t \sum_{j=1}^4 e_j(t) \; i_j (t) \; dt
\end{equation}
Using equations (\ref{current}) and (\ref{voltage}), the energy expression becomes:
\begin{equation} \label{Energy}
\begin{aligned}
            E(t) = \int_0^t \sum_{j=1}^4[c_0 + c_1 \; \omega_j(t) + c_2 \; \omega_j^2(t) + c_3 \; \omega_j^3(t) \\
    + c_4 \; \omega_j^4(t) + c_5 \; \dot{\omega_j}^2(t)]dt
\end{aligned}
\end{equation}

where,
\begin{subequations} \label{constants}
\small
    \begin{align}
        & c_0 = \frac{RT_f^2}{K_T^2} , \; \; c_1 = \frac{T_f}{K_T}\left(\frac{2RD_f}{K_T} + K_E\right)\\
        & c_2 = \frac{D_f}{K_T}\left(\frac{RD_f}{K_T} + K_E\right) + \frac{2RT_fk_{\tau}}{K_T^2} \\
        &c_3 = \frac{k_{\tau}}{K_T}\left(\frac{2RD_f}{K_T} + K_E\right),  \; \;c_4 = \frac{Rk_{\tau}^2}{K_T^2}, \; \; c_5 = \frac{2RJk_{\tau}}{K_T^2}
    \end{align}
\end{subequations}

\begin{figure*}
    \begin{minipage}{0.73\textwidth}
         \includegraphics[width =\textwidth]
   {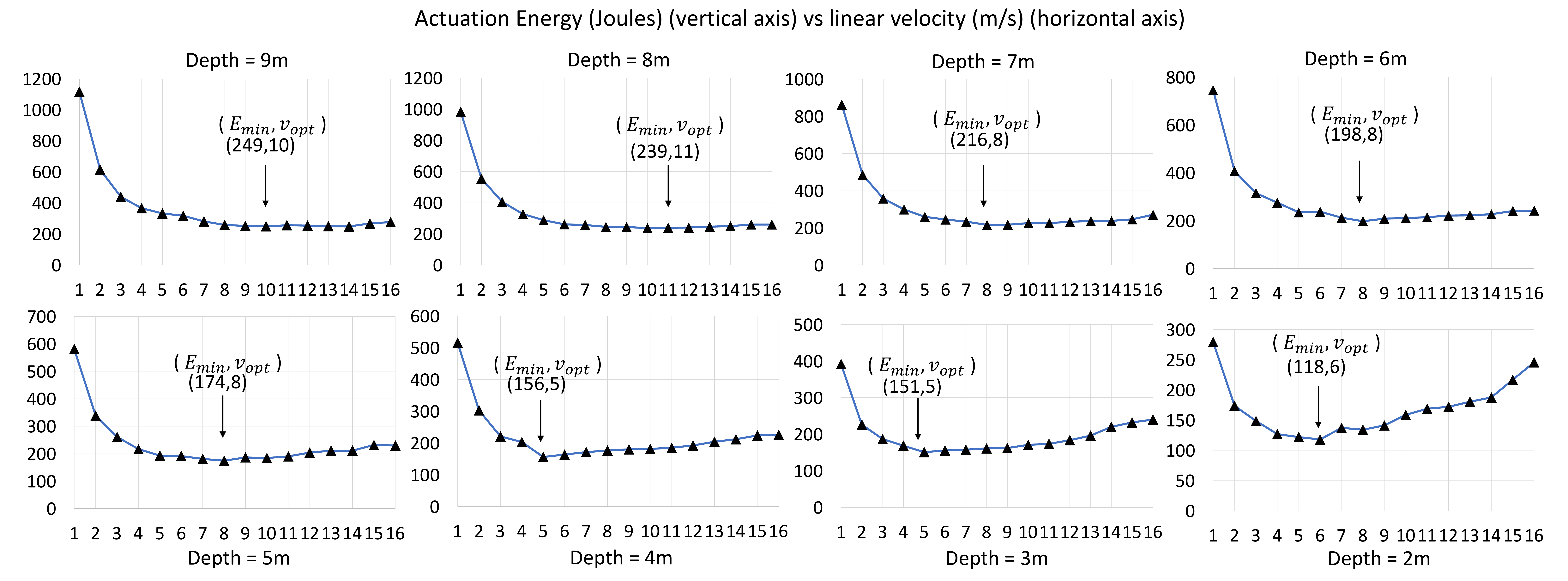} 
   \vspace{-7mm}
    \caption{Energy burned by quadrotor motors for traversing different depths for varying times of flight. \\This only considers linear motion along one direction.}
    \label{fig:en}
    \end{minipage}
    \begin{minipage}{0.25\textwidth}
 \includegraphics[width = \textwidth]   {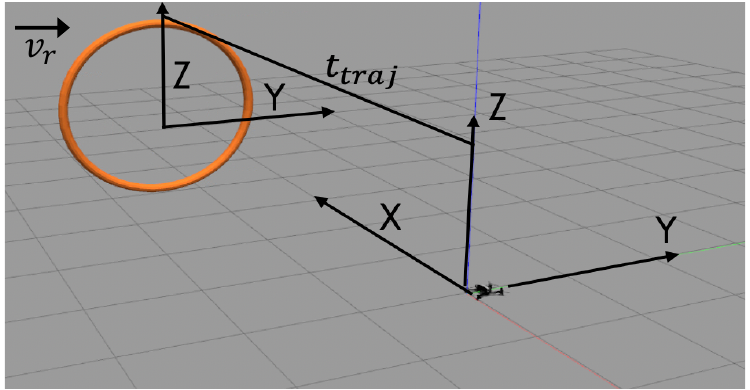} 
    \caption{Gazebo world containing the dynamic gate moving with velocity $v_r$ which changes direction after reaching distance $+/- L$. $t_{traj}$ is the trajectory time  estimated by the PgNN.}
    \label{fig:ring}
    \end{minipage}
    \vspace{-6mm}
\end{figure*}

Using the expression given in equation (\ref{Energy}), and plugging in the constant values given in equations (\ref{constants}), we are thus able to calculate the energy for a particular trajectory up to time $t$.  Given a certain distance away from a target destination (depth), an energy-efficient planner should be able to approximately estimate the flight time that would minimize the energy.  We utilize the knowledge obtained through the aforementioned equations by simulating several trajectories to reach targets situated at different depths.  

\subsubsection{\textbf{Energy characterization as a function of velocity}}
\label{physim}
We implemented the energy model using the values in Table \ref{tab:en}. We present key insights into our physics-guided simulation. Figure \ref{fig:en} presents the actuation energy for traversing different depths. Using our physics-guided simulation, we report an average instantaneous power of $\sim124W$ for a $2700mAh$ battery. For each depth, we swipe the linear velocity as the control variable from $1$ m/s to $16$ m/s (the top speed of the drone in this study) in steps of $1$ m/s. We observe that the energy is the highest when the robot speed is the minimum. The time taken by the drone to reach the destination varies inversely with speed. As a consequence, if we use Eqn. (\ref{Energy}) to estimate the energy, then the effect of time dominates at a lower speed for which the energy adds up. With increasing speed, the energy burned drops dramatically in the beginning due to a reduction in flight time. Beyond a certain speed, the energy again increases. This is because, for higher speeds, the motor currents drawn from the battery increases, resulting in an increase in instantaneous power and consequently energy. We observe there is a sweet spot for a near-optimal velocity which yields the least energy for a particular trajectory. 
This establishes the need for an interpolation  a near-optimal velocity, given the target depth.


\subsubsection{\textbf{Physics-Guided Neural Network}}
\label{pgnn_train}
We collect several trajectories as part of our training dataset by flying the drone across different depths and by varying the linear velocity as shown in Figure \ref{fig:en}. A vanilla ANN can be trained with several such depth-velocity pairs. However, the ANN will have no knowledge about the underlying source data distribution.

\begin{subequations}\label{profile}
\begin{align}
        E = c_0 + c_1v + c_2v^2 + c_3v^3 + c_4v^4 + c_5v^5 \\
        \frac{dE}{dv} = c_1 + 2c_2v + 3c_3v^2 + 4c_4v^3 + 5c_5v^4
\end{align}
\end{subequations}

As shown in Equation (\ref{profile}a), we fit 5th-order polynomial curves to the energy-velocity profile obtained from the simulations, for each depth. To approximately evaluate the optimal velocity for near-minimum energy, we equate the derivative of the fitted polynomial to zero as shown in Eqn. (\ref{profile}b). {As shown in Table \mbox{\ref{tab:data}}, for each training data point, the expression from Eqn. \mbox{(\ref{profile}b)} using the corresponding polynomial coefficients is added as an additional regularization term in the loss function. While a mid-range value of $v$ will cover most cases, addition of physical equations makes the model more interpretable.}
We use the standard mean-square error loss along with the additional physical constraints:
\begin{equation}
    \mathcal{L} = \frac{1}{n}\sum_{i=1}^{n}||v_i^* - v_i^{pred}||^2 + \lambda \sum_{i=1}^{n}\sum_{j=1}^{5}jc_jv_i^{*(j-1)}
\end{equation}
$n$ is the number of training samples. $v^*$ is the desired velocity obtained from the groundtruth, while $v^{pred}$ is the velocity the network learns. $\lambda$ is the regularization coefficient.

\begin{table}[t]
\setlength\belowcaptionskip{-20pt}
\centering
\tiny
\caption{Dataset for Physics-guided Neural Network}
\label{tab:data}
\renewcommand{\arraystretch}{0.95}
\resizebox{\columnwidth}{!}{%
\begin{tabular}{c|c|c}
\hline
\footnotesize{\textbf{Depth}} & \footnotesize{\textbf{Velocity}} & \footnotesize{\textbf{Constraint}}                                                 \\ \hline 
\footnotesize{$d_1$}          & \footnotesize{$v_1$}             & \footnotesize{$c_{11} + 2c_{21}v_1 + 3c_{31}v_1^2 + 4c_{41}v_1^3 + 5c_{51}v_1^4 = 0$} \\ 
\footnotesize{$d_2$}          & \footnotesize{$v_2$}             & \footnotesize{$c_{12} + 2c_{22}v_2 + 3c_{32}v_2^2 + 4c_{42}v_2^3 + 5c_{52}v_2^4 $ = 0}\\ 
.....          & ....              & ....                                                                \\ 
\footnotesize{$d_n$}          & \footnotesize{$v_n$}             & \footnotesize{$c_{1n} + 2c_{2n}v_n + 3c_{3n}v_n^2 + 4c_{4n}v_n^3 + 5c_{5n}v_n^4 $ = 0}\\ \hline
\end{tabular}%
}
\end{table}
\raggedbottom
\vspace{-2mm}
\begin{equation} \label{ttraj}
    t_{traj} = \frac{d}{v^{pred}}
\end{equation}

Note, the output of the PgNN is a velocity. Using Eqn. (\ref{ttraj}), we calculate an approximate  trajectory time using which we estimate the position of the moving gate, in the future. Also, the training set is small because we are constraining the training via physical equations. This enhances sample-efficiency which accelerates training. While the training set consists of a similar range of depths as seen during inference, the inference can have a depth value not available in training data, making the proposed method superior than a regression-less lookup-based approach.
\vspace{-4mm}
\subsection{\textbf{Planning Robot Trajectories}}
\label{symbol}
\subsubsection{\textbf{Motion Estimation of Moving Gate}} As shown in Figure \ref{fig:ring}, we assume the moving gate has a linear velocity along the Y axis as $v_r$. We use our SNN-based object tracking to estimate $v_r$. Consider the position of the gate along the Y axis to be $y_1$ at any given instant, obtained from the object tracker. For a sensing interval of $\delta t$, let the position of the same point in the gate after $\delta t$ be $y_2$. Then, to estimate the instantaneous velocity of the gate, we use the expression given:
\begin{equation}
    v_r = \frac{y_2-y_1}{\delta t}
\end{equation}

\begin{algorithm}[!h]
\label{algo:planner}
\setstretch{0.75}
\small
\SetAlgoLined
 \textbf{Require:}  $t_{traj}$, $L$ , $y_1$, $y_2$\tcp{\footnotesize{Input Arguments}}
 $v_r \gets \frac{y_2 - y_1}{\delta t}$\tcp{\footnotesize{$v_r$ estimate via SNN}}
  $d_1 \gets v_r \times t_{traj}$ \tcp{\footnotesize{Gate distance traversed}} 
  Compute distance   $d_2 = f(L,y_2)$ from approaching end \\
  
  \begin{multicols}{2} 
  \If{ $y_2 > y_1$}{
    \tcp{\footnotesize{Gate moving right}} 
  \If{$y_2 > 0$}
    {$d_2 = L - y_2$\\}
  \If{$y_2 < 0$}
    {$d_2 = L + |y_2|$\\}
   \footnotesize{Compare $d_1$ and $d_2$}\\
   \If{$d_1 > d_2$ \tcp{\footnotesize{direction change > $d_2$}}} 
    {$y^* = L - d_1 + d_2$\\} 
  \If{$d_1 < d_2$}
    {$y^* = y_2 + d_1$\\}  
 }
 \columnbreak
   \If{ $y_2 < y_1$}{
    \tcp{\footnotesize{Gate moving left}} 
  \If{$y_2 < 0$}
    {$d_2 = |-L - y_2|$\\}
  \If{$y_2 > 0$}
    {$d_2 = L + y_2$\\}
   \footnotesize{Compare $d_1$ and $d_2$}\\
   \If{$d_1 > d_2$ \tcp{\footnotesize{direction change > $d_2$}}}
    {$y^* = -L + d_1 - d_2$\\}
  \If{$d_1 < d_2$}
    {$y^* = y_2 - d_1$\\}  
 }
  \end{multicols}
 return $y^*$
 \caption{$trajectory\_learner.py$}
\end{algorithm}
\raggedbottom

\subsubsection{\textbf{Symbolic Planning}}
\label{symb_plan}
We assume the gate is constrained within distances $+L$ and $-L$, beyond which the gate reverses the direction of motion. This is the only prior knowledge that we use during inference. Algorithm 
\ref{algo:planner} presents the pseudo-code of our symbolic predictive planner. {The planner accepts the variables $t_{traj}$ (trajectory time) and $L$ as the input arguments along with $y_1$ and $y_2$. $y_1$, $y_2$ are the outputs from the SNN and $t_{traj}$ is the output from the PgNN. The information from the two networks are now combined as discussed next.} The algorithm outputs $y^*$ -- the desired destination of the robot after $t_{traj}$, which will enable collision-free gate crossing. The distance $d_1$ traversed by the gate is computed during the trajectory time $t_{traj}$ (Line 3). Also,  the distance $d_2$ (which is the distance of the gate from the nearest end ($-L$ or $+L$) ) is computed by considering the different sub-cases. If the gate moves right (line 5), then we consider the cases where the gate is in the positive Y axis (Line 6), or negative Y axis (Line 9). Accordingly, the value of $d_2$ is evaluated (Line 7 or Line 10). 
If $d_1$ $>$ $d_2$, then the gate will change the direction of motion (from right to left), and $y^*$ is computed (Line 15). If $d_1$ $<$ $d_2$, then the gate keeps moving right, and $y^*$ is computed accordingly (Line 18).
Similarly, if the gate is detected to be moving left (Line 21), $y^*$ is evaluated symbolically by stepping through the hard-coded rules (Lines 22 - 36). { Once $y^*$ is obtained, we use the minimum jerk trajectory \mbox{\cite{mellinger2011minimum}} which is optimized to make sure the amount of current drawn from the battery is the lowest possible which results in almost zero rate of change of acceleration. Consequently, the planned trajectory burns near-minimum motor currents for the concerned navigation task.
Note, that during PgNN training, the predicted velocity is along the depth axis (path AB = depth $d$ in Figure \mbox{\ref{fig:near_min}}). However, during inference, there will be a component of the drone velocity along the Y axis ($v_y$ in Figure \mbox{\ref{fig:near_min}) in the same direction as $v_r$}, as the target destination is C which depends on the speed of the moving ring ($v_r$ which cannot be known in advance). The predicted drone velocity $v^{pred}$ is only a proxy to calculate $t_{traj}$ (see Eqn. \mbox{\ref{ttraj}}). However, a lower value of $t_{traj}$ will reduce the difference between paths AB and AC, making the actuation-energy near-minimum. Also, note, the flight path has a parabolic nature, but for the sake of explanation we simplified them as straight lines in Figure \mbox{\ref{fig:near_min}}. While it is beyond the scope of this work to mathematically derive lower bounds, our method is reasonably energy-efficient (see Sections \mbox{\ref{actuation_en}}, \mbox{\ref{ablation}} and \mbox{\ref{compare}).} }


\vspace{-3mm}

\begin{table}[!h]
\centering
\tiny
\caption{Neural Network Specifications}
\vspace{-2mm}
\label{tab:impl}
\renewcommand{\arraystretch}{0.8}
\footnotesize
\resizebox{\columnwidth}{!}{%
\begin{tabular}{cc|cc}
\hline
\multicolumn{2}{c|}{\textbf{SNN   Parameters}}     & \multicolumn{2}{c}{\textbf{Physics-guided   ANN Parameters}}        \\ \hline
\multicolumn{1}{c|}{Kernel   (W)  Size}      & $3\times3$  & \multicolumn{1}{c|}{Number of Layers}  & $3$                \\ 
\multicolumn{1}{c|}{W{[}0,0,1,1{]}}                 & $0.15$ & \multicolumn{1}{c|}{Neurons per layer} & {[}$64,128,128${]} \\ 
\multicolumn{1}{c|}{Leak Factor   ($\beta$)} & $0.1$ & \multicolumn{1}{c|}{ Coefficient ($\lambda$)} & $1e-04$ \\ 
\multicolumn{1}{c|}{ Threshold ($U_{th}$)} & $1.75$ & \multicolumn{1}{c|}{Epochs}            & $2000$             \\ \hline
\end{tabular}%
}
\end{table}
\raggedbottom

\section{Methodology}
\label{method}
We implemented the SNN for event-based tracking using SNNTorch \cite{snntorch}, whereas the physics-guided ANN was implemented using Tensorflow \cite{tensorflow2015-whitepaper}. For the PgNN, we also used batch normalization after each layer. The moving gate was created using blender \cite{blender}. To integrate the proposed planner with a controller, we used the RotorS simulator \cite{rotorS} and spawned the drone using the Gazebo physics engine \cite{gazebo}. Table \ref{tab:impl} mentions the parameters used for the networks.
\vspace{-2mm}

\section{Experimental Results}
\label{res}
As a comparative scheme, we implemented an equivalent fusion-less physics-guided planner.  It uses a depth camera to track the gate as well. We call this the \emph{Depth-Planner}. 
\subsection{\textbf{Object Tracking}}
\label{track}
Figure \ref{iou} illustrates the mean and peak intersection-over-union (IOU) values for the object tracking using event-based SNN for different depths. The corresponding visual representations are shown in Figure \ref{fig:evbox}. For depth $2$m, the IOU is lesser than depths $3$m or $4$m (where we observe best tracking). At a depth less than $2$m, the moving gate (which has a diameter of $2$m) comes too near to fit within the $640 \times 480$ viewing window (see Figure \ref{fig:evbox}). For depth greater than $6$m, the performance degrades. The success rate of the overall design reduces with increased perception overhead. Hence, we only report performance statistics up to a depth of $6$m subsequently.

\begin{figure}[!t]
\begin{center}
   \includegraphics[width = 0.38\textwidth]
   {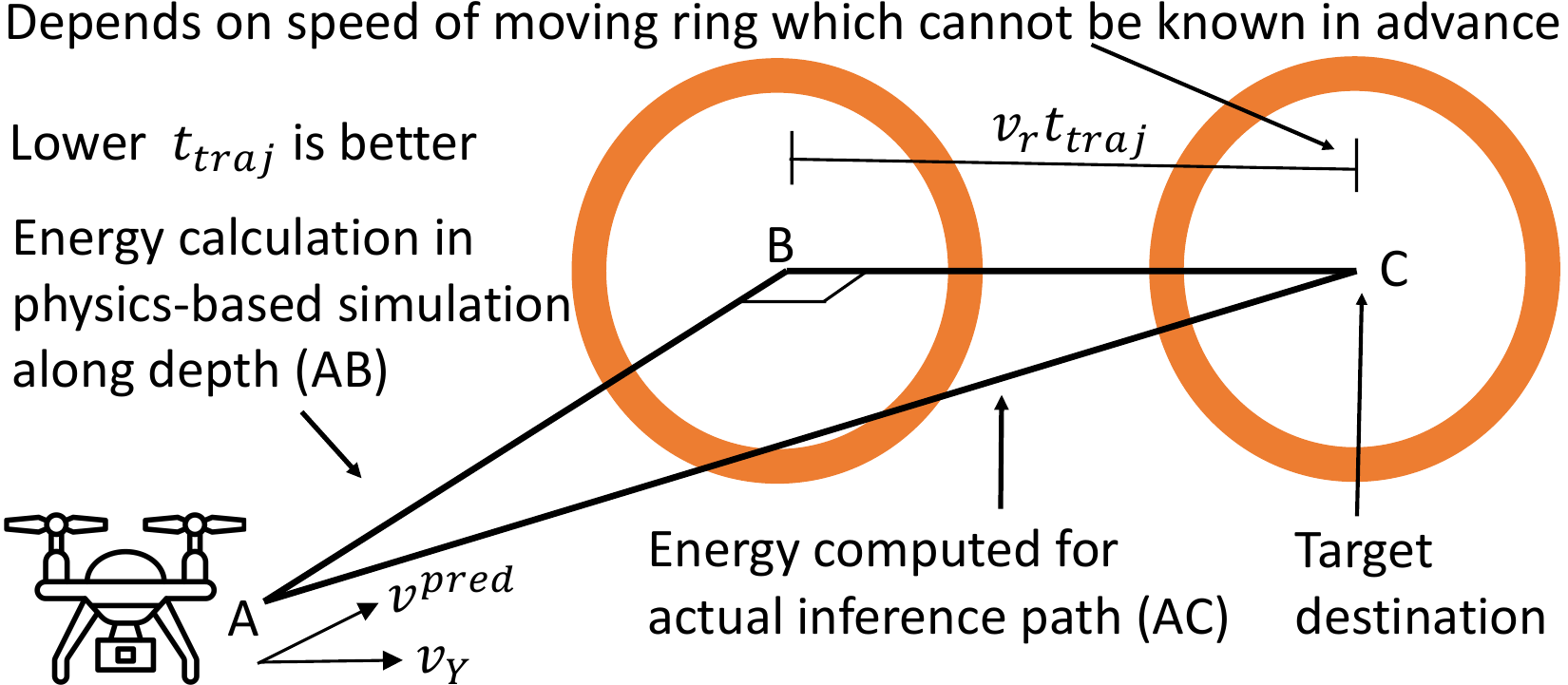} 
\end{center}
\vspace{-3mm}
    \caption{{Actuation Energy: Training vs Inference}}
    \vspace{-4mm}
    \label{fig:near_min}
\end{figure}
\vspace{-3mm}
\subsection{\textbf{Success Rate}}
\label{success_rate}
\begin{figure*}[!t]
\vspace{1.5mm}
    \begin{minipage}{0.33\textwidth}
         \includegraphics[width =\textwidth]
   {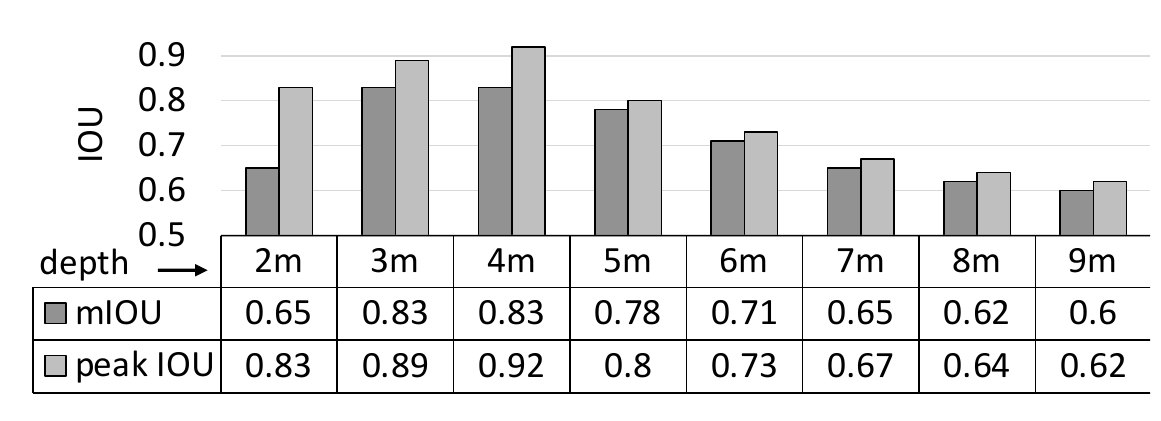} 
   \vspace{-6mm}
   \caption{Object Tracking using Event-based SNN }
   \label{iou}
    \end{minipage}
    \begin{minipage}{0.65\textwidth}
 \includegraphics[width = \textwidth]{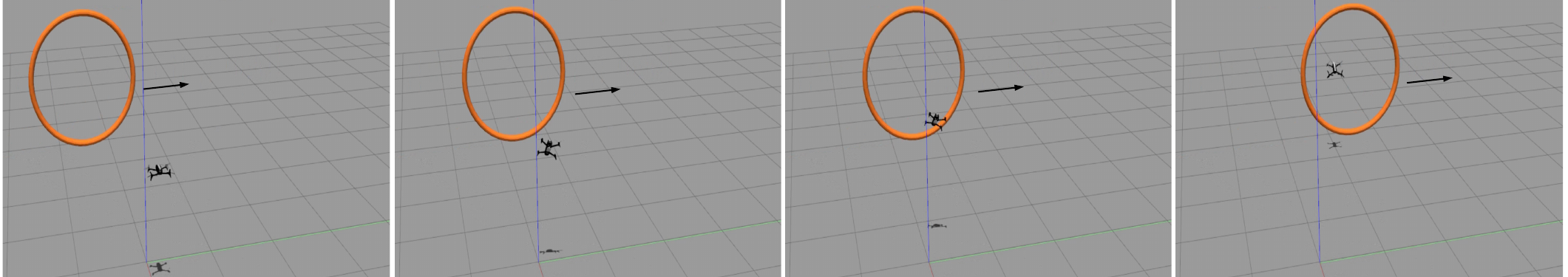} 
 \caption{Parrot Bebop2 crossing a moving gate using the \emph{EV-Planner} algorithm.}
 \label{traj}
    \end{minipage}
    \vspace{-4mm}
\end{figure*}

Figure \ref{traj} illustrates in a time-series snapshot the {Parrot Bebop2} drone flying through the moving gate using \emph{EV-Planner}. Table \ref{tab:sr} enumerates the success rate of collision-free gate crossing for different starting points of both the drone and the gate. 
Hence, the results showcase the generalization capability of our design for points not present in the training data. The success rates are lower when the drone starts from $y = \pm2$ as  the gates are detected (entirely) for lesser duration (especially at depths $< 3$m). For certain cases, the sensor-fused \emph{EV-Planner} performs better than fusion-less \emph{Depth-Planner}. {Tracking an object using depth involves a series of intermediate image processing steps that incur significant latency. By the time the target is evaluated, the gate shifts by a  little, which \mbox{\emph{Depth-Planner}} struggles to take into account (for greater depths mostly).} As the depth becomes greater, this effect is observed more for both methods (IOU reduces in Figure \ref{iou}).  \emph{EV-Planner} still performs better, as event-based tracking is faster than depth-based tracking.
\begin{table*}[]
\centering
\caption{Success Rates for different initial locations. Each data-point considers 10 observations. Higher is better.}
\vspace{-2mm}
\label{tab:sr}
\small
\renewcommand{\arraystretch}{0.9}
\resizebox{1.8\columnwidth}{!}{%
\begin{tabular}{cc|ccc|ccc|ccc|ccc|ccc}
\hline
\multicolumn{2}{c|}{\textbf{Depth}} &
  \multicolumn{3}{c|}{\textbf{2m}} &
  \multicolumn{3}{c|}{\textbf{3m}} &
  \multicolumn{3}{c|}{\textbf{4m}} &
  \multicolumn{3}{c|}{\textbf{5m}} &
  \multicolumn{3}{c}{\textbf{6m}} \\ \hline
\multicolumn{1}{c|}{\multirow{2}{*}{\textbf{\begin{tabular}[c]{@{}c@{}}Starting \\ Points\end{tabular}}}} &
  Drone (x,y) &
  \multicolumn{1}{c}{(0,0)} &
  \multicolumn{1}{c}{(0,$\pm$1)} &
  (0,$\pm$2) &
  \multicolumn{1}{c}{(1,0)} &
  \multicolumn{1}{c}{(1,$\pm$1)} &
  (1,$\pm$2) &
  \multicolumn{1}{c}{(2,0)} &
  \multicolumn{1}{c}{(2,$\pm$1)} &
  (2,$\pm$2) &
  \multicolumn{1}{c}{(3,0)} &
  \multicolumn{1}{c}{(3,$\pm$1)} &
  (3,$\pm$2) &
  \multicolumn{1}{c}{(4,0)} &
  \multicolumn{1}{c}{(4,$\pm$1)} &
  (4,$\pm$2) \\ \cline{2-17} 
\multicolumn{1}{c|}{} &
  Ring (-2,y) &
  \multicolumn{1}{c}{$\pm$2} &
  \multicolumn{1}{c}{$\mp$1} &
  $\pm$1 &
  \multicolumn{1}{c}{$\pm$2} &
  \multicolumn{1}{c}{$\mp$1} &
  $\pm$1 &
  \multicolumn{1}{c}{$\pm$2} &
  \multicolumn{1}{c}{$\mp$1} &
  $\pm$1 &
  \multicolumn{1}{c}{0} &
  \multicolumn{1}{c}{0/$\pm$1} &
  0 &
  \multicolumn{1}{c}{$\pm$2} &
  \multicolumn{1}{c}{$\pm$1/$\mp$2} &
  $\mp$2 \\ \hline
\multicolumn{1}{c|}{\multirow{2}{*}{\textbf{\begin{tabular}[c]{@{}c@{}}Success  \\  Rate\end{tabular}}}} &
  \begin{tabular}[c]{@{}c@{}}Depth-Planner\\ \end{tabular} &
  \multicolumn{1}{c}{0.8} &
  \multicolumn{1}{c}{0.7} &
  0.7 &
  \multicolumn{1}{c}{1} &
  \multicolumn{1}{c}{0.9} &
  0.8 &
  \multicolumn{1}{c}{1} &
  \multicolumn{1}{c}{0.9} &
  0.8 &
  \multicolumn{1}{c}{0.9} &
  \multicolumn{1}{c}{0.8} &
  0.7 &
  \multicolumn{1}{c}{0.7} &
  \multicolumn{1}{c}{0.6} &
  0.5 \\ \cline{2-17} 
\multicolumn{1}{c|}{} &
  \begin{tabular}[c]{@{}c@{}}EV-Planner\\ \end{tabular} &
  \multicolumn{1}{c}{1} &
  \multicolumn{1}{c}{0.8} &
  0.8 &
  \multicolumn{1}{c}{1} &
  \multicolumn{1}{c}{0.9} &
  0.9 &
  \multicolumn{1}{c}{1} &
  \multicolumn{1}{c}{1} &
  0.9 &
  \multicolumn{1}{c}{0.9} &
  \multicolumn{1}{c}{0.9} &
  0.8 &
  \multicolumn{1}{c}{0.7} &
  \multicolumn{1}{c}{0.7} &
  0.6 \\ \hline
\end{tabular}%
}
\end{table*}
\raggedbottom

\begin{figure*}
\vspace{-3mm}
    \begin{minipage}{0.59\textwidth}
         \includegraphics[width =\textwidth]
   {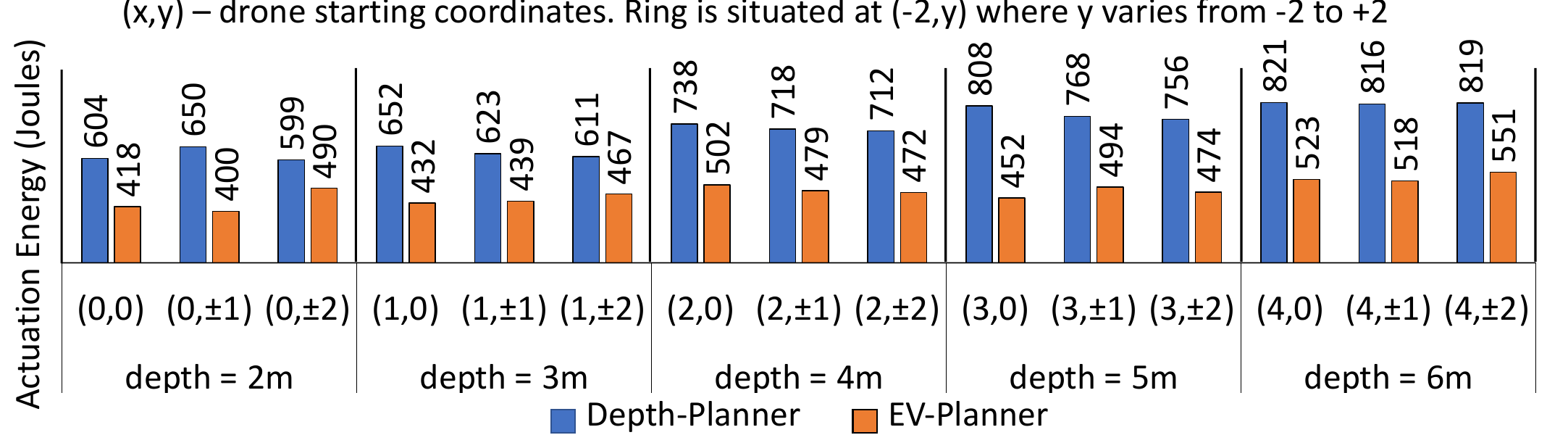} 
   \vspace{-6mm}
   \caption{Actuation Energy (averaged over 25 flights with each flight averaged over 10 runs) for different starting points of the drone while flying through the ring without colliding. \\ Lower is better. }
   \label{fig:evp_en}
    \end{minipage}
    \begin{minipage}{0.4\textwidth}
 \includegraphics[width = \textwidth]{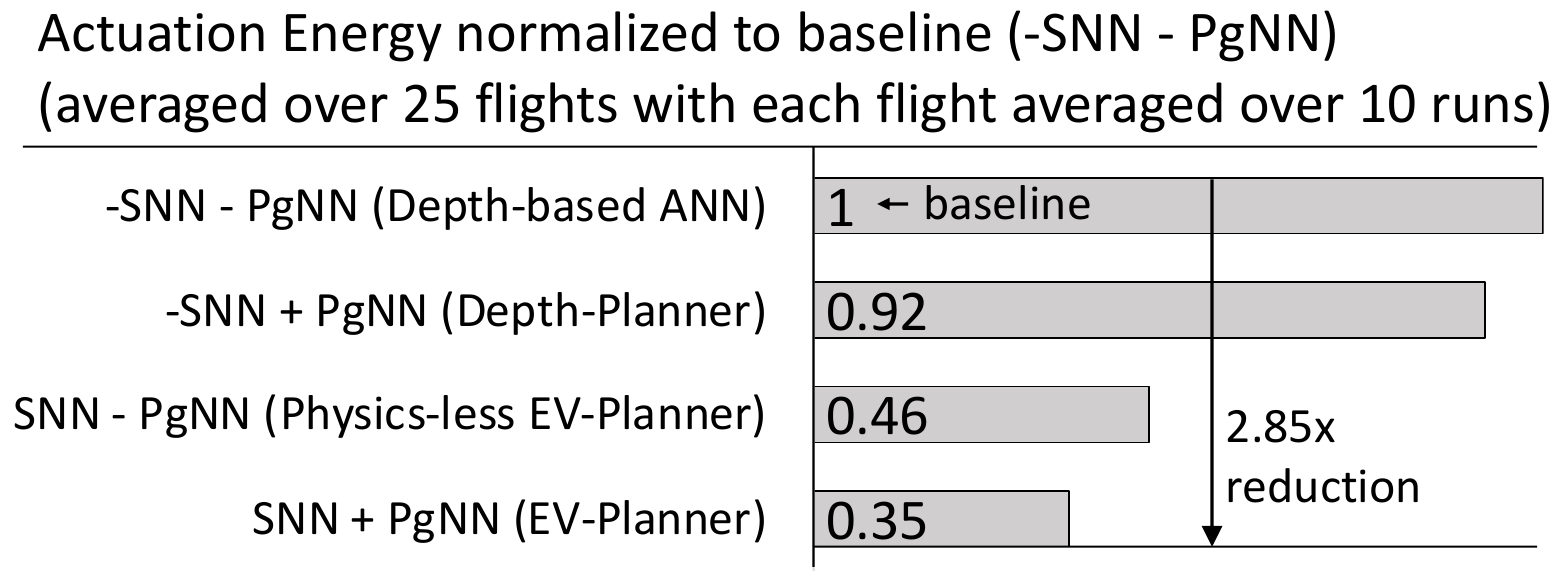} 
 \vspace{-7mm}
 \caption{{Ablation Study for actuation energy to quantify \\ the contribution of each design component. Lower is better.}}
 \label{fig:abl}
    \end{minipage}
    \vspace{-4mm}
\end{figure*}

\subsection{\textbf{Actuation Energy: Event vs Depth}}
\label{actuation_en}
Figure \ref{fig:evp_en} presents the average actuation energy for \emph{EV-Planner}
compared to \emph{Depth-Planner} for different starting locations of the drone. 
Although we do not compare computation energy for perception, the increased latency for depth-based tracking increases the actuation energy. This is because, the drone hovers longer for the depth-based tracking, which in turn burns more rotor currents. 
{On average,  \mbox{\emph{Depth-Planner}} takes $2$ seconds longer for the symbolic planner.  For an average instantaneous power of $124W$, this translates to an additional $248J$  of energy, as depth-based perception overhead. On the other hand, \mbox{\emph{EV-Planner}} relies on events for faster object tracking using low-latency SNNs. For $25$ flights (each averaged over 10 runs), \mbox{\emph{EV-Planner}} achieves $2.6\times$ lower actuation energy compared to \mbox{\emph{Depth-Planner}}.}
\subsection{\textbf{Ablation Study}}
\label{ablation}
{
Figure  \mbox{\ref{fig:abl}} presents an ablation study to quantify the contribution of each design component (i.e the SNN based perception and PgNN based flight-time estimator). Note that when SNN is not used, the depth-sensor based tracking is employed (\emph{Depth-Planner}). In absence of the PgNN, a vanilla ANN without any underlying physical prior is used for time-estimation. A systematic comparison among the four possible cases reveal that SNN + PgNN based \emph{EV-Planner} results in $\sim 2.85 \times$ lower actuation energy compared to the baseline which uses depth-based perception and physics-less planning, highlighting the energy-efficiency of our proposed design.}
\vspace{-3mm}

\subsection{\textbf{Comparison with SOTA Event-based Perception}}
\label{compare}
{\mbox{Table \ref{tab:params}} presents a summary by comparing two state-of-the-art (SOTA) perception works which use events for object-detection -- a) Recurrent Vision Transformers (RVT) \mbox{\cite{_EV3}} and b) Asynchronous Event-based Graph Neural Networks (AEGNN) \mbox{\cite{_EV4}}. \mbox{\emph{EV-Planner}} which uses a lightweight variant of \mbox{\cite{nagaraj2023dotie}} is scene-independent without requiring any training and is also orders of magnitude more parameter-efficient.
RVT and AEGNN both detect objects elegantly, but require curated datasets with lots of instances and labelling. AEGNN uses different configurations for different datasets. To use any supervised-learning (SL) technique in place of our shallow SNN, one has to prepare a dataset with several instances of the ring along with proper labels resulting in training overhead. Also, since the model complexity is greater than the shallow SNN, we did not use SL-based methods as perception block.
}

\begin{table}[!h]
\tiny
\caption{{Model Complexity Comparison for event-based perception}}
\label{tab:params}
\resizebox{\columnwidth}{!}{%
\begin{tabular}{c|c|c}
\hline
\textbf{Works} & \textbf{Architecture Details} & \textbf{Mode of Learning} \\ \hline
\cite{_EV3}       & \begin{tabular}[c]{@{}c@{}}4 transformer blocks with\\  $\sim$9-18 million parameters\end{tabular} & Supervised   \\ \hline
\cite{_EV4}       & \begin{tabular}[c]{@{}c@{}}7 graph convolution blocks, 2 pooling layers\\ $\sim$230k parameters for ncars dataset\end{tabular}   & Supervised   \\ \hline
This work & \begin{tabular}[c]{@{}c@{}}1 3x3 Conv2d block with 9 spiking neurons \\ with no trainable parameters\end{tabular}                & Unsupervised \\ \hline
\end{tabular}%
}
\vspace{-4mm}
\end{table}

\section{Limitations, Discussions and Future Work}
\label{discuss}
 The results clearly highlight the benefits associated with sensor-fused (events and depth) perception for navigation via SNNs and PgNNs. For future evaluations in real-world, noise could be an issue for which the perception pipeline needs modification. While event cameras are more robust to low-light and motion blur\mbox{\cite{evsurvey}}, it is possible to encounter spurious events in the environment. One way to filter those spurious events would be to introduce clustering \cite{nagaraj2023dotie}. However, that will add to the latency overhead which may be unavoidable in real-world experiments but not encountered in a simulator.
 Also, note that, though the PgNN predicted velocities would result in actuation energies (as low as $\sim200J$), the planner burns more power during the perception window for gate tracking and  due to the Y component of the velocity as shown in Figure \mbox{\ref{fig:near_min}}. Hence, the energies reported for the actual navigation are higher. There have been works that evaluate compute energy for object detection by considering special-purpose hardware \cite{reram, Roy_2023_CVPR}. {Compute or non-actuation energy refers to the perception energy which is typically executed on low-power processors with  current ratings of nano-Amperes. The actuators are the rotor blades which draws currents of several Amperes ($10^9\times$ more w.r.t compute current) from a 15 V battery, making the actuation energy orders of magnitude higher. Hence, we limit our energy calculations in this work only to actuation energy}. Research on reducing compute energy will consider designing embedded sensor-fused controllers with scaled in-robot compute hardware with programming support in future. This will reduce the payload mass, and hence the actuation energy. Future work will also consider deploying the proposed method in a real robot with the aforementioned considerations.

\section{Summary}
  We presented \emph{EV-Planner} -- an event-based physics-guided neuromorphic planner to perform obstacle avoidance using neuromorphic event cameras and physics-based AI. 
For object tracking, we used event-based low-latency SNNs.
Utilizing the physical equations of the  drone rotors, an energy-aware PgNN was trained with depth inputs to predict flight times. The outputs from the neural networks were consumed by a symbolic planner to publish the robot trajectories. The task of
autonomous quadrotor navigation was considered with the aim to detect
moving gates and fly through them while avoiding a collision.
Extensive simulation results show that sensor-fusion based \emph{EV-Planner} performs  generalizable collision-free dynamic gate crossing with decent success rates, with {lower actuation energy compared to ablated variants.}


%





\ifCLASSOPTIONcaptionsoff
  \newpage
\fi




\bibliographystyle{./bibliography/IEEEtran}
\bibliography{./bibliography/main}

\end{document}